\documentclass[conference]{IEEEtran}  
\usepackage{natbib}  
\usepackage{amsmath}    % For mathematical equations
\usepackage{graphicx}   % For including figures 
\usepackage{hyperref}   % For hyperlinks
\usepackage{amssymb}    % For additional symbols
\usepackage{algorithm}  % For algorithms
\usepackage{algpseudocode} % For algorithmic pseudocode
\usepackage{booktabs}
\usepackage{xcolor}
\usepackage{multirow}
\usepackage{subcaption} 
\usepackage{colortbl}
\usepackage{paralist}
\definecolor{highlight}{HTML}{BFFFBF}
\newcommand{\remove}[1]{}

\title{ 
Predicting Time Series of Networked Dynamical Systems without Knowing Topology
} 
\author{ 
    Yanna Ding\textsuperscript{\rm 1}, 
    Zijie Huang\textsuperscript{\rm 2}, 
    Malik Magdon-Ismail\textsuperscript{\rm 1}, 
    Jianxi Gao\textsuperscript{\rm 1} \\
    \textsuperscript{\rm 1}Rensselaer Polytechnic Institute\\
    \textsuperscript{\rm 2}University of California, Los Angeles \\
    \texttt{\{dingy6, magdon, gaoj8\}@rpi.edu}, \texttt{zijiehuang@cs.ucla.edu}
}

\begin{document}

\maketitle

\begin{abstract}
Many real-world complex systems, such as epidemic spreading networks and ecosystems, can be modeled as networked dynamical systems that produce multivariate time series. Learning the intrinsic dynamics from observational data is pivotal for forecasting system behaviors and making informed decisions. However, existing methods for modeling networked time series often assume known topologies, whereas real-world networks are typically incomplete or inaccurate, with missing or spurious links that hinder precise predictions. Moreover, while networked time series often originate from diverse topologies, the ability of models to generalize across topologies has not been systematically evaluated. To address these gaps, we propose a novel framework for learning network dynamics directly from observed time-series data, when prior knowledge of graph topology or governing dynamical equations is absent. Our approach leverages continuous graph neural networks with an attention mechanism to construct a latent topology, enabling accurate reconstruction of future trajectories for network states. Extensive experiments on real and synthetic networks demonstrate that our model not only captures dynamics effectively without topology knowledge but also generalizes to unseen time series originating from diverse topologies.
\end{abstract}
 
\section{Introduction}
\begin{figure*}[t]
  \centering
  \includegraphics[width=\linewidth]{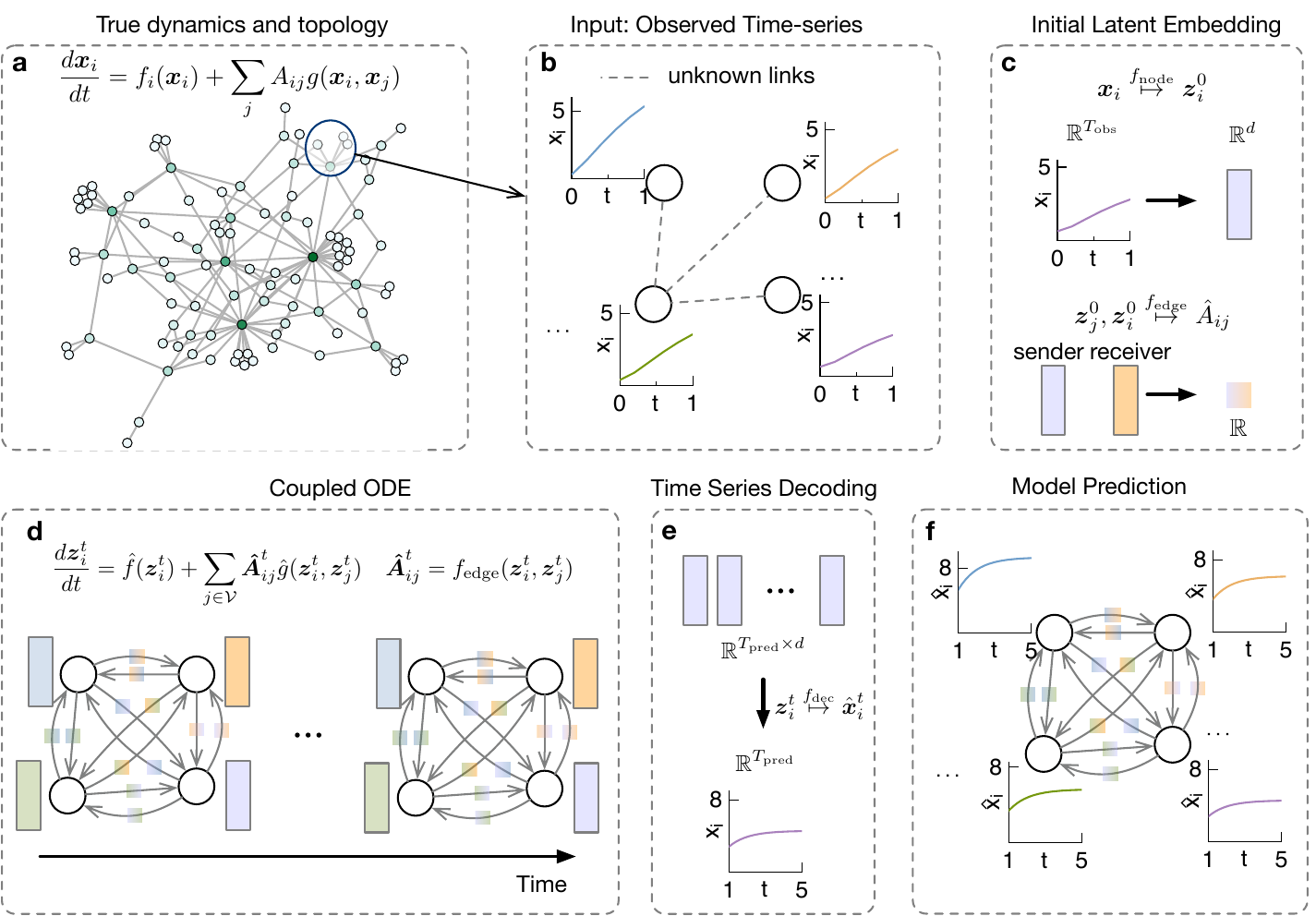}
  \caption{Problem setup and Model Illustration. \textbf{a}, Ground truth nodal ODE $\frac{d\boldsymbol{x}_i}{dt}$ and topology that govern the network states. In the ODE, $f_i, g$ denote the self-dynamics and interaction term respectively and adjacency matrix $\boldsymbol{A}$ describes the network structure. Nodes with a higher degree have a more intense colors. \textbf{b}, The model takes an initial period of time-series data for each node as input, with the edges assumed to be unknown. \textbf{c}, The encoder maps each nodal trajectory to a corresponding hidden vector and infers network interactions based on the nodal embeddings. \textbf{d}, The neural ODE  drives the latent state evolution.   \textbf{e}, The hidden representation at each timestamp is decoded to the input space. \textbf{f}, The model forecasts nodal trajectories beyond the observation window.  \label{fig:demo} } 
\end{figure*}
Dynamical network systems are pervasive across various domains, including epidemic spreading networks, where nodal states reflect the number of infected individuals~\citep{pastor2001epidemic}, population flow networks capturing geospatial population dynamics~\citep{gardiner1985handbook}, and gene regulatory networks modeling the interactions between genes and their expression products~\citep{alon2006introduction}. 
Learning dynamical systems is essential for solving various problems, particularly in predicting future system states~\citep{gao2022autonomous,wu2023predicting,huang2024treat,CFGODE}, but also in controlling networks and analyzing transitions in behavior~\citep{liu2011controllability,morone2019k}. In this paper, we focus on the reverse engineering of networked dynamical systems from time-series data, with the goal of accurate time-series prediction.

Existing approaches to modeling complex system dynamics often assume access to the true network topology. However, in practice, the ground truth network structures are often concealed, and observed structures may contain missing or erroneous links~\citep{von2002comparative,guimera2009missing,timme2014revealing,wang2016data}. For instance, when modeling the spread of diseases, one might consider administrative divisions like states as nodes and construct the network based on factors such as population movements or airline traffic. However, the measurement of these movements  may contain inaccuracies~\citep{barbosa2018human}, and the assumptions used to construct the network might not fully capture the complexities of the actual topology. Consequently, developing methods to predict future network trajectories without explicit knowledge of network topology has become a key research area~\citep{prasse2022predicting}.

Another underexplored aspect pertains to the generalizability~\citep{yehudai2021local} of models to unseen topologies in the context of networked time-series forecasting. Existing works either train and test models on the same networked time series, referred to as the transductive setting~\citep{huang2020learning,zang2020neural,tran2021radflow}, or on different sets of networked time series, known as the inductive setting~\citep{kipf2018neural,huang2020learning,yin2021leads}. Methods in the transductive setting often overlook the variability in network structures that arises in some real-world systems, where networks representing the same underlying dynamics—such as a gene cohort at different developmental stages—may differ in topology~\citep{alon2006introduction}. Although methods in the inductive setting consider different topologies, they lack systematic evaluation across diverse network types, such as random, scale-free, or community networks, particularly under out-of-distribution (OOD) conditions involving unseen topologies. Addressing these gaps is critical for understanding the robustness and adaptability of networked time-series forecasting models in diverse and evolving network environments.  

We propose a novel neural network model to address the challenges of networked time-series forecasting in the absence of prior topology information. Our model extracts relational patterns among network components from an initial short period of observed time-series data and employs neural Ordinary Differential Equations (ODEs)~\citep{chen2018neural} to predict nodal trajectories at future timestamps. Each node is mapped to a latent representation, allowing the neural ODE to model the evolution of these hidden embeddings over time. By effectively modeling the relationships between network components through latent topologies that potentially evolve over time, our approach captures the underlying network dynamics and flexibly adapts to both static systems with fixed topologies and dynamic environments where network structures change chronologically.

Our contributions are summarized as follows:
\begin{enumerate}
    \item Dynamic network modeling without topology knowledge: We introduce a novel model for learning networked dynamics without requiring predefined topological structures. By leveraging latent representations and neural ODEs, our model effectively captures dynamic interactions from time-series data, providing a flexible and robust solution for a wide range of applications.
    \item Comprehensive validation: We evaluate our model on both transductive and inductive settings using real and synthetic datasets. Our experiments span a variety of dynamical systems, including epidemic spreading models~\citep{pastor2001epidemic,prasse2022predicting}, population dynamics~\citep{gardiner1985handbook,novozhilov2006biological}, gene regulatory networks~\citep{alon2006introduction,karlebach2008modelling}, ecological dynamics~\citep{macarthur1970species,gao2016universal}, and neural activity dynamics~\citep{wilson1972excitatory,laurence2019spectral}. The results demonstrate competitive performance across various dynamics compared to baseline methods.
    \item Systematic evaluation across network models: We asses model performance across diverse network types, addressing OOD scenarios where training and testing datasets are drawn from different topological distributions.  
\end{enumerate}

%\vspace{-1em}
 
\section{Related Work}
\textbf{Networked Time-Series Prediction.}
The goal of network time-series prediction is to forecast nodal states over time.
Symbolic regression approaches~\citep{bongard2007automated, schmidt2009distilling, brunton2016discovering, gao2022autonomous} aim to uncover explicit dynamical equations from observed data. These methods often require predefined functional forms and struggle to generalize to complex, unknown systems.
Graph-based neural ODE models, such as LGODE~\citep{huang2020learning}, CGODE~\citep{huang2021coupled}, and HOPE~\citep{luo2023hope}, extend neural ODEs by incorporating graph neural networks to capture spatiotemporal dependencies. Neural Relational Inference (NRI)~\citep{kipf2018neural} employs a variational encoder-decoder architecture that models pairwise interactions and state transitions in a discrete setting.  Despite their flexibility, these models often rely on known or partially known network structures. In contrast, our work eliminates the need for prior topology knowledge, offering a robust solution for forecasting network dynamics and generalizing across diverse and unseen network structures.

\textbf{Network Inference.}
Another line of research focuses on network reconstruction~\citep{de2010advantages,peixoto2019network}, which aims to infer network topology from observed time-series data to reveal the structural interactions within the system. Regression-based approaches include \citep{casadiego2017model}, which estimates interaction strengths using basis function expansions, and \citep{prasse2018network, prasse2020network}, which infer adjacency weights by treating them as parameters relying on known dynamical formulas. \cite{zhang2022universal} combines a deep learning approach with Markovian assumptions to jointly infer adjacency matrices and dynamical parameters, focusing on network reconstruction and prediction in a discrete framework.  Unlike these approaches, our work bypasses explicit network reconstruction and instead uses latent relationships to predict nodal trajectories.

\textbf{Learnability of Networked Dynamical Systems.}
The PAC framework has been employed to study the learnability of discrete networked dynamical systems, providing theoretical insights into network interactions and node functions.  \cite{narasimhan2015learnability} investigates the PAC learnability of influence functions in partially observed social networks, while \cite{adiga2019pac} focuses on threshold functions within known network topologies.  \cite{qiu2024learning} further examines the learnability of node functions in networked dynamical systems under partial topology knowledge. Our work complements them by tackling the practical challenge of forecasting nodal dynamics without prior knowledge of topology or dynamical rules.

\section{Method}
\subsection{Preliminaries}

Consider a graph denoted as $\mathcal{G}=(\mathcal{V},\mathcal{E})$ with nodes $\mathcal{V}=\{1,\ldots,N\}$ and edges $\mathcal{E}\subset \mathcal{V} \times \mathcal{V}$. 
The nodal activity observed at the $\tau$th time step ($\tau\in \mathbb{Z}_{>0}$)  is a $D$-dimensional vector $\boldsymbol{x}_i^{\tau}\in \mathbb{R}^{D}$. This activity can describe the probability of infection in a pathology spreading network~\citep{jiang2020true}  or the velocity and location in charged particle systems~\citep{zhang2022universal}.  
Suppose each node's time series has $T$ timestamps in total. 
The temporal nodal states can be arranged as:
$
\boldsymbol{x}_{i} = [\boldsymbol{x}_{i}^1,\ldots,\boldsymbol{x}_{i}^{T }]$  ($i\in \{1,\ldots, N\}$).
We use $\boldsymbol{x}_i^{t_1:t_2}$ to denote a subsequence of $\boldsymbol{x}_i$ including all observations from timestamp $t_1$ to $t_2$ ($t_1\leq t_2$): $\boldsymbol{x}_i^{t_1:t_2} = [\boldsymbol{x}_i^{t_1},\boldsymbol{x}_i^{t_1+1},\ldots,\boldsymbol{x}_i^{t_2}]$. 
The network connections and interacting strength are compactly represented as the $N\times N$ adjacency matrix $\boldsymbol{A}$. We assume the coupling strength $\boldsymbol{A}_{ij}$ is a real number, with a larger value indicating a stronger impact from node $j$ to node $i$. 
While the observation is sampled at discrete time, the underlying dynamics is continuous. We denote the state of $i$ at a continuous time $t\in\mathbb{R}$ as $\boldsymbol{x}_i(t)$.
Each nodal trajectory is governed by a coupled ordinary differential equation~\citep{gao2016universal}
\begin{align}
\frac{d \boldsymbol{x}_i(t)}{dt} 
&= \mathcal{F}\left(f(\boldsymbol{x}_i(t)) + \sum_j \boldsymbol{A}_{ij}  g(\boldsymbol{x}_i(t),\boldsymbol{x}_j(t))\right)\label{eq:ode}
\end{align}
Given any initial condition, Integrating Eq~\eqref{eq:ode} produces a sequence of states for $i$.
Here $f:\mathbb{R}^{D}\rightarrow \mathbb{R}^D, g(\cdot,\cdot):\mathbb{R}^{D}\times \mathbb{R}^D \rightarrow \mathbb{R}^D$ describe the self-dynamics and the interaction term between adjacent nodes, respectively. Function $\mathcal{F}$ combines the self-feedback and neighborhood impact.  

We assume $\boldsymbol{A}_{ij} \geq 0$ and consider two types of interaction terms: $\frac{\partial g}{\partial \boldsymbol{x}_j}>0$ and $\frac{\partial g}{\partial \boldsymbol{x}_j} <0$ to reflect whether the system is entirely cooperative or competing. We use the notation that a vector is greater than zero if all of its entries are greater than zero. 

We define the dynamics learning model as follows: our model takes as input an initial short period of observations on each node, denoted as $\boldsymbol{x}_i^{1:T_{\text{obs}}}$, where $T_{\text{obs}}$ is the length of the condition window. The output of our model is the prediction of the states for each individual at subsequent time steps $T_{\text{obs}}+1, T_{\text{obs}}+2, \ldots, T $, represented as $\boldsymbol{x}_i^{T_{\text{obs}}+1:T }$.  We  refer to the time interval $[T_{\text{obs}}+1,T ]$ as the prediction window.
%\vspace{-1em}
\subsection{Model Definition}  

In this paper, we present TAGODE (Topology-Agnostic Graph ODE) for
learning networked dynamical systems without prior knowledge of the underlying network structure.
Our model consists of three primary modules: an encoder, a neural ODE, and a decoder. The encoder's role is to deduce the initial conditions of latent vectors ($\boldsymbol{z}_i(0)$) for individual nodes.  The decision on whether the encoder produces the edge embedding depends on the specific type of neural ODE. We consider two variations of neural ODE: one with evolving latent edges and another without. In the first case,  the encoder solely produces node embeddings, and the neural ODE captures the interaction at each specific time instant $t$. In the latter case, the encoder creates a latent edge embedding. As we solve an initial value problem using the neural ODE with the initial condition $\boldsymbol{z}_i(0)$, it results in a sequence of latent states, spanning $T -T_{\text{obs}}$ time steps for each node. Subsequently, the decoder independently maps each latent state back to the input space, generating predictions for nodal states at time steps $T_{\text{obs}}+1$ through $T$.  The terms ``FeedForward network'' and ``Multi-layer Perceptron'' are used interchangeably in our context.

\textbf{Encoder.}
The encoder module for nodes, represented by $f_{\text{node}}$, takes a segment of observations for each node's activity $\boldsymbol{x}_i^{1:T_{\text{obs}}}\in\mathbb{R}^{T_{\text{obs}}D}$ as input and maps them to a latent initial condition of dimension $d$. The role of the encoder is dependent on whether the neural ODE assumes a fixed edge embedding. In the case where a fixed edge embedding is employed, the encoder's responsibility extends to generating representations of pairwise interactions. These interactions between nodes are computed based on the concatenation of their initial latent states. This process can be mathematically expressed as follows: 
\begin{align}
\boldsymbol{z}_i(0)&= f_{\text{node}}(\boldsymbol{x}_i^{1:T_{\text{obs}}}) \\
\boldsymbol{\hat{A}}_{ij} &= f_{\text{edge}} ([\boldsymbol{z}_j(0)|| \boldsymbol{z}_i(0)])
\end{align}
where $\boldsymbol{z}_i(0)$ represents the latent initial condition, $\boldsymbol{\hat{A}}_{ij} \in \mathbb{R}$ is the effect of latent node $j$ on $i$, and $||$ denotes vector concatenation.

This architecture allows flexibility in the choice of $f_{\text{node}}$, which can be implemented as a Feedforward network (FFW), a graph transformer~\citep{yun2019graph}, an NRI-based encoder~\citep{kipf2018neural} or a spatiotemporal graph representation encoder~\citep{luo2023hope,huang2020learning,huang2021coupled}. The latter two encoders require knowledge of the network topology, so we assume a fully connected underlying structure in such cases. For a detailed comparative study of these encoder variants, please refer to Appendix~\ref{sec:exp_encoder}.

For the remaining sections, we will assume the adoption of the Feed-forward module to define the encoder functions $f_{\text{node}},f_{\text{edge}}$.
In our experiments, we utilized the following specific instance of the Feedforward module:
\begin{align}
\mathrm{FeedForward}(\boldsymbol{h}) &= \boldsymbol{W}^{(2)}
\sigma(\boldsymbol{W}^{(1)}\boldsymbol{h}  + \boldsymbol{b}^{(1)}) +\boldsymbol{b}^{(2)}\label{eq:ffw}
\end{align}
In Eq~\eqref{eq:ffw},   $\boldsymbol{W}^{(l)} \in \mathbb{R}^{d\times \cdot}, \boldsymbol{b}^{(l)} \in\mathbb{R}^{d  }$ are learnable weights and biases and $\sigma$ is any nonlinear activation function such as ReLU or GELU~\citep{hendrycks2016gaussian}.
We compute latent nodes independently because we do not assume prior knowledge of topology. This approach allows each observation to contribute to the node's latent vector without any predefined network structure.

\textbf{Neural ODE and Decoder.}
We define the Ordinary Differential Equation (ODE) governing the evolution of latent states in our model. To capture the self-dynamics and interactions between latent nodes, we adopt a universal framework, which expresses the dynamics as follows:
\begin{align} 
\frac{d\boldsymbol{z}_i(t)}{dt} &= \phi_i(\boldsymbol{z}_1(t),\ldots,\boldsymbol{z}_N(t), {\boldsymbol{\hat{A}}}) \\
&=\hat{\mathcal{F}}\left(\hat{f}(\boldsymbol{z}_i(t)) + \sum_{j=1}^{N} \boldsymbol{\hat{A}}_{ij} \hat{g}(\boldsymbol{z}_i(t) ,\boldsymbol{z}_j(t))\right) \label{eq:node1}
\end{align}
While \cite{casadiego2017model} employed a linear combination of basis functions to model the ODE of observable nodal states $\boldsymbol{x}_i$, where the accuracy of their approach relied on selecting appropriate basis functions, our objective is to model the evolution of latent states. These latent states can exhibit more complex functional forms. Therefore, we choose to parameterize the dynamical formulas $\hat{f} $ and $\hat{g} $ using FeedForward networks, allowing us to learn these functions directly from the data.
The latent state at each timestamp can be computed using the following integral formulation:
\begin{align} 
\boldsymbol{z}_i^\tau &= \int_{t=0}^{\tau} \phi_i(\boldsymbol{z}_1(t),\ldots,\boldsymbol{z}_{N} (t), {\boldsymbol{\hat{A}}}) \ dt
\end{align}

We further consider a model variant (TAGODE-VE) incorporating  time-varying edges to take into consideration potential evolution of interaction.  Unlike the first type of ODE~\eqref{eq:node1}, where the encoder provides constant edge weights, TAGODE-VE leverages a multi-head attention mechanism to infer pairwise interactions. This mechanism computes attention scores for all pairs of nodal embeddings, enabling the model to adapt and capture changing interactions as they unfold over time.
The attention scores, represented as  $ {\boldsymbol{\hat{A}}}_{ij}^{h}$ are obtained through a softmax function applied to learned attention logits $\boldsymbol{e}_{ij}^h$. These logits are computed using key and query matrices, $\boldsymbol{W}_k^h$ and $\boldsymbol{W}_q^h$, which project the sender and receiver nodes, respectively. The LeakyReLU activation function is employed to introduce non-linearity and enhance the model's ability to capture complex interactions. The formula for the   score of one attention head is given as follows:
\begin{align}
{\boldsymbol{\hat{A}}}_{ij}^{h} &= \text{softmax}(\boldsymbol{e}_{ij}^h)\\
\boldsymbol{e}_{ij}^h &= \text{LeakyReLU}((\boldsymbol{W}_k^{h} \boldsymbol{z}_j)^{\top} (\boldsymbol{W}_q^{h} \boldsymbol{z}_i)).
\end{align} 
We aggregate the outputs from $n_{\text{head}}$ attention heads to approximate the interaction term. The corresponding latent nodal ODE is defined as follows:
\begin{align}
\frac{d\boldsymbol{z}_i (t)}{dt} &=\hat{\mathcal{F}}\left(
\hat{f}(\boldsymbol{z}_i(t)) +  \sum_{j=1}^{N} \overset{n_{\text{head}}}{\underset{h=1}{\parallel}} 
 \boldsymbol{\hat{A}}_{ij}^{h}\hat{g}(\boldsymbol{z}_i(t) ,\boldsymbol{W}_v^h\boldsymbol{z}_j(t))\right) \label{eq:node2}
\end{align}
The value matrix $\boldsymbol{W}_v^h$ transforms the expression of the sender node to extract the feature from $j$ that has an impact on $i$.
By aggregating these individual heads, our model gains a more comprehensive understanding of the underlying interactions.
As demonstrated in the experimental section~\ref{sec:ood}, both ODE types perform similarly when inferring in-distribution dynamics. However, the attention-based ODE exhibits superior generalization capabilities when handling out-of-distribution dynamics.

The model performs decoding by processing the hidden vectors individually at each timestamp to generate state predictions:
\begin{align}
\hat{\boldsymbol{x}}_i^\tau&=f_{\text{dec}}(\boldsymbol{z}_i^\tau).  
\end{align}
In our experiment setup, $f_{\text{dec}}:\mathbb{R}^{d} \rightarrow \mathbb{R}^{D}$ is implemented as a Multi-layer Perceptron (MLP).
We define the objective function as the state reconstruction loss over the prediction window:
\begin{align}
\text{minimize } \sum_{i=1}^{N}\sum_{\tau =T_{\text{obs}}+1}^{T } 
\lVert \hat{\boldsymbol{x}}_i^\tau - \boldsymbol{x}_i^\tau\rVert_2^2  . 
\end{align}

We formulate the edge embedding modeling within an unsupervised framework. Our goal is to deduce a latent topology that enables accurate predictions of network states, rather than explicitly recovering the ground truth network structure. This approach proves beneficial in scenarios where the topology information is deprecated or incomplete.

\begin{table*}[ht!]
  \centering
  \scriptsize
  \caption{ Transductive learning evaluation on COVID-19 dataset with 1-week, 2-weeks, and 3-weeks-ahead prediction length over 5 runs.  Our model variant with time-evolving edge weights has a trailing ``VE'' indicating time-varying edge embeddings.    \label{tab:covid} } 
  \resizebox{\textwidth}{!}{%
  \begin{tabular}{lcccccccccccc}
    \toprule
    \multirow{2}{*}{Method} & \multicolumn{2}{c}{1-week-ahead}  &   \multicolumn{2}{c}{2-weeks-ahead} & \multicolumn{2}{c}{3-weeks-ahead} \\
    \cmidrule(lr){2-3} \cmidrule(lr){4-5} \cmidrule(lr){6-7}
      & MAE & MAPE & MAE & MAPE & MAE & MAPE \\
    \midrule
    LSTM      &   446.042$\pm$334.427 &  .159$\pm.$089 & 993.233$\pm$603.299 & .337$\pm.$163 & 1694.365$\pm$1081.436 & .526$\pm.$231 \\
GRU       &   299.316$\pm$299.551 &  .110$\pm.$071 & 675.437$\pm$649.614 & .251$\pm.$160 & 1156.715$\pm$1139.533 & .399$\pm.$308 \\
NRI         & 128.236$\pm$20.525  &  .043$\pm.$020 & 300.777$\pm$162.759 & .122$\pm.$094 & 942.714$\pm$1272.727  & .689$\pm.$960 \\
AIDD        & 163.389$\pm$27.725  &  .034$\pm.$007 & 398.799$\pm$11.186  & .074$\pm.$013 & 515.274$\pm$225.757   & .103$\pm.$020 \\
NODE        & 108.814$\pm$6.224   &  .026$\pm.$002 & 210.891$\pm$7.644   & .050$\pm.$003 & 244.602$\pm$42.227    & .067$\pm.$006 \\
\cmidrule{1-7}
TAGODE        & 100.031$\pm$3.717   &  \textbf{.025$\pm.$001} & \textbf{183.402$\pm$10.232}  & \textbf{.045$\pm.$004} &  {197.989$\pm$18.643}    & \textbf{.046$\pm.$005} \\
TAGODE-VE   & \textbf{97.115$\pm$5.080}    &  \textbf{.025$\pm.$001} & 191.486$\pm$12.172  & .046$\pm.$003 & \textbf{185.723$\pm$ 12.962}   & .048$\pm.$003\\
    \bottomrule
  \end{tabular}}
   
\end{table*}
 
\section{Experiments}
In this section, we empirically evaluate our model in a transductive learning setting with real epidemic spreading data (Section~\ref{sec:transductive}) and an inductive learning setting using synthetic data generated from diverse dynamical models (Section~\ref{sec:inductive}). Furthermore, we examine its performance under OOD testing (Section~\ref{sec:ood}). Additional experiments, including analyses of scalability, robustness, and ablation studies, are detailed in Appendix~\ref{sec:additiona_exp}.
\subsection{Experiment Setup }

\textbf{Baselines.} We compare with four autoregressive baselines and a neural ODE-based method.
\begin{inparaenum}[(i)]
    \item \textbf{LSTM}: The LSTM baseline adopted by \citep{kipf2018neural}. It models time-series independently for each node. A 2-layer MLP is added both before and after an LSTM unit, which outputs a one-step lookahead value for the hidden state.  
    \item \textbf{GRU}~\citep{cho2014learning}: Gated Recurrent Unit. Similar to the LSTM baseline, it includes 2-layer MLPs to perform feature transformation, but the one-step lookahead is predicted by a GRU unit.
    \item \textbf{NRI}~\citep{kipf2018neural}: The Neural Relational Inference model. NRI performs message passing between nodes and edges to produce the probability of the system state at the next timestamp.  
    \item \textbf{AIDD}~\citep{zhang2022universal}. The Automated Interactions and Dynamics Discovery (AIDD) model parameterizes the weighted adjacency matrix and dynamical equations for Markov dynamics. 
    \item \textbf{NODE}~\citep{chen2018neural}: Neural Ordinary Differential Equations. The vanilla NODE    can be regarded as only modeling the self-dynamics for each node, without consideration of interaction. In our implementation, we employ the same encoder and decoder for NODE as in our model and utilize a FeedForward network to parameterize the self-dynamics. 
\end{inparaenum}    

For autoregressive models, prior to the observation cutoff at $T_{\text{obs}}$, the input state is the observation. After $T_{\text{obs}}$, the model's own predictions at previous time steps serve as input. The model size of LSTM, GRU, and AIDD scale with the number of nodes, while ours is independent of the network size.    Our implementation is publicly available\footnote{\url{https://github.com/dingyanna/LatentTopoDynamics}}.

\textbf{Evaluation Metric.} We use Mean Absolute Percentage Error (MAPE), Mean Absolute Error (MAE), and Root Mean Squared Error (RMSE) to evaluate the model performance, with their formulas detailed in Section~\ref{sec:appendix_metric}.

\subsection{Transductive Setting Evaluation}\label{sec:transductive}
In this section, we focus on a transductive learning setting by applying a chronological train/test split on a single multivariate time-series. In this case, both train and test data are sampled from an identical dynamical system with the same underlying topology. We evaluate models on a real dataset  \textbf{COVID-19}~\citep{dong2020interactive}, which consists of the state-level reported cases of coronavirus in the United States. We utilize seven  features to form the nodal state, including five dynamic features collected from the Johns Hopkins University (JHU) Center for Systems Science and Engineering (\#Confirmed, \#Deaths, \#Recovered, Testing-rate, Mortality-rate), a static feature of state population, and a in-state population flow value provided by SafeGraph2\footnote{\url{https://www.safegraph.com/covid-19-data-consortium}}. The model is trained on data from April.12.2020 to Nov.30.2020 and tested on the time span between Dec.01.2020 and Dec.31.2020. We condition on an observation window of length 21, and predict the cumulative  deaths in the future 1-week, 2-weeks, and 3-weeks. Please refer to Appendix~\ref{appendix_covid} for more details.  

The results on this dataset is summarized in Table~\ref{tab:covid}. TAGODE demonstrates a reduction in MAE (MAPE) compared to the best-performing non-continuous baseline by 24.27\% (26.47\%), 39.02\% (39.19\%), and 61.58\% (55.34\%) for the three specified prediction ranges.
In comparison to the continuous baseline NODE, our models incorporate the modeling of latent interactions. This additional feature results in an average decrease in MAE (MAPE) by 37.54\% (20.06\%) and 14.71\% (13.40\%) for TAGODE and TAGODE-VE, respectively. This improvement is most pronounced in the longest prediction task, demonstrating the superior performance of our models for relatively long-term predictions.
The findings also indicate that continuous models excel in learning the dynamics of multi-agent systems in real-world applications.

\subsection{Inductive Setting Evaluation\label{sec:inductive}}
In this section, we systematically evaluate model generalizability under two conditions:
\begin{inparaenum}[(i)]
    \item testing within known topological distributions, where training and testing datasets share the same topology type  but differ in specific wiring; and
    \item  testing under  OOD  conditions, where both the degree   and edge weights distributions are entirely unseen during training.  
\end{inparaenum} 

\textbf{Dynamics.}
We study six types of continuous dynamics, as explored in~\citep{macarthur1970species,hens2019spatiotemporal,prasse2022predicting}: \begin{inparaenum}[(i)]
\item \textbf{SIS}: Epidemic spreading, modeled using the Susceptible-Infected-Susceptible (SIS) framework; 
\item \textbf{Population}: Population dynamics driven by birth-death processes; 
\item \textbf{Regulatory}: Gene regulatory networks; 
\item \textbf{Mutualistic}: Mutualistic plant-pollinator interactions; 
\item \textbf{Neural}: Neuronal activity; 
\item \textbf{Lotka-Volterra}: Predator-prey systems, modeled using the Lotka-Volterra equation.
\end{inparaenum}
The ODE formulas corresponding to these dynamics are summarized in Table~\ref{tab:dyn_formula}.  The training and testing trajectories are generated by solving the ODE with a specified initial condition. 
 
\begin{table}[h]
    \centering
    \scriptsize 
        \caption{Dynamical Formulas. Unspecified dynamical parameters are sampled from a predetermined distribution, given in Appendix~\ref{appendix_synthetic_data}, Table~\ref{tab:ic}.   }
    \resizebox{\columnwidth}{!}{%
    \begin{tabular}{lllll  } 
    \toprule
    Dynamics & $f_i(\boldsymbol{x}_i)$  & $g(\boldsymbol{x}_i,\boldsymbol{x}_j)$
    \\    
    \toprule 
    SIS & $-\delta_i \boldsymbol{x}_i$ & $(1-\boldsymbol{x}_i ) \boldsymbol{x}_j $ \\ 
    Population & $- \boldsymbol{x}_i^{0.5} $ & $\boldsymbol{x}_j^{0.2}$ \\ 
    Regulatory & $-\boldsymbol{x}_i$ & $\boldsymbol{x}_j^2(1+\boldsymbol{x}_j^2)^{-1}$\\  
    Mutualistic  & $  \boldsymbol{x}_i (1- {\boldsymbol{x}_i^{2}} )$ & $\boldsymbol{x}_i\boldsymbol{x}_j (1+\boldsymbol{x}_j)^{-1}$  \\ 
    Neural & $-\boldsymbol{x}_i$ & $(1+\exp(-\tau(\boldsymbol{x}_j -\mu)))^{-1}$  \\  
    Lotka-Volterra &$\boldsymbol{x}_i (\alpha_i - \theta_i \boldsymbol{x}_i)$   & $-\boldsymbol{x}_i\boldsymbol{x}_j$  \\
    \bottomrule
    \end{tabular} }
    \label{tab:dyn_formula}
\end{table}  
 
\textbf{Network Topology.}
We explore three types of synthetic network models: Erdős-Rényi (\textbf{ER}) networks with Poisson-distributed node degrees, scale-free  (\textbf{SF})  networks with a power-law degree distribution generated using preferential attachment, and \textbf{Community} networks generated via random partition graphs. For each dynamics paired with a topology type, we generate 140 network realizations with 100 nodes, and report results for both individual and mixed datasets.
More details about these dynamics, topology types and their simulation are deferred to Appendix~\ref{appendix_synthetic_data}.

\begin{table*}[ht]
\scriptsize
\centering
\caption{Inductive learning evaluation on synthetic networks. For Lotka-Volterra, RMSE is reported and MAPE is reported for the rest of the dynamics.  Our models include TAGODE and TAGODE-VE. The latter has time-varying edges. \label{tab:comp}}
\resizebox{\textwidth}{!}{%
\begin{tabular}{clcccccc}
 \toprule
& Method& SIS & Population & Regulatory  & Mutualistic  & Neural  & Lotka-Volterra \\
\midrule
\multirow{7}{*}{ER} &
LSTM & 23.3137$\pm$1.4676&.2527$\pm.$0258&.1565$\pm.$0119&.2104$\pm.$0215&.1743$\pm.$0136&.3812$\pm.$0247\\
& GRU &1.5448$\pm.$1526&.2786$\pm.$0322&.1987$\pm.$0122&1.0196$\pm.$1008&.1870$\pm.$0145&9.4544$\pm.$0130\\
& NRI &4.7384$\pm.$2785&.7990$\pm.$0862&.7037$\pm.$0136&\textbf{.1151$\pm.$0216}&.6169$\pm.$0098&1.1010$\pm.$0052 \\
& AIDD &.2168$\pm.$0178&13.7869$\pm.$6797&.3237$\pm.$0234&.4064$\pm.$4821&.2773$\pm.$0199&.3236$\pm.$0459\\ 
& NODE & .0123$\pm.$0085&.0187$\pm.$0017&\textbf{.0239$\pm.$0035}&.2754$\pm.$0656&.0262$\pm.$0041&.1342$\pm.$0242\\ 
\cmidrule{2-8}
& TAGODE   &.0180$\pm.$0093&\textbf{.0164$\pm.$0020}&.0244$\pm.$0037&.1644$\pm.$2247&.0552$\pm.$0049&.1157$\pm.$0195
 \\
& TAGODE-VE &\textbf{.0098$\pm.$0065}&.0166$\pm.$0022&.0267$\pm.$0044& {.1411$\pm.$1825}&\textbf{.0225$\pm.$0044}&\textbf{.0954$\pm.$0224} \\
\midrule
\multirow{7}{*}{SF} 
& LSTM & 96.3243$\pm$33.7797&.2217$\pm.$0390&.1825$\pm.$0150&.2642$\pm.$1637&.2168$\pm.$0141&.2849$\pm.$0254\\
& GRU  & 2.3759$\pm.$9383&2510$\pm.$0435&.7273$\pm.$0379&2.5366$\pm$2.9231&.3650$\pm.$0301&.5054$\pm.$0762 \\
& NRI & .8572$\pm.$4381&.4178$\pm.$0455&.2337$\pm.$0110&\textbf{.1928$\pm.$2376}&.2139$\pm.$0156&.3468$\pm.$0309\\
& AIDD &.5880$\pm.$2128&.6206$\pm.$0285&.3214$\pm.$0304&.8205$\pm$1.5992&.3650$\pm.$0180&.6356$\pm.$0546 \\ 
& NODE & .1884$\pm.$1935&.0295$\pm.$0021&.0385$\pm.$0030&.3364$\pm.$1918&.0316$\pm.$0031&.2456$\pm.$0217\\ 
\cmidrule{2-8}
& TAGODE  &.1413$\pm.$1196&\textbf{.0275$\pm.$0017}&\textbf{.0323$\pm.$0027}& {.3174$\pm.$2275}&\textbf{.0269$\pm.$0044}& {.1291$\pm.$0229} \\
& TAGODE-VE & \textbf{.0783$\pm.$0283}&.0275$\pm.$0039&.0396$\pm.$0037&.3227$\pm.$3201&.0276$\pm.$0035&\textbf{.1264$\pm.$0235} \\
\midrule
\multirow{7}{*}{Community} 
& LSTM & 14.0208$\pm$1.1742&.2018$\pm.$0268&.1592$\pm.$0072&.1772$\pm.$0089&.1410$\pm.$0112&.4942$\pm.$0096\\
& GRU & 3.9153$\pm.$3393&.2124$\pm.$0273&.1861$\pm.$0089&.2964$\pm.$0228&.1511$\pm.$0120&2.0445$\pm.$0161  \\ 
& NRI & .1889$\pm.$0198&.9515$\pm.$0019&.1527$\pm.$0120&.0981$\pm.$0432&.6723$\pm.$0113&.2681$\pm.$0396 \\
& AIDD & .2714$\pm.$0354&1.5232$\pm.$1503&.2554$\pm.$0188&.2371$\pm.$1773&.2414$\pm.$0200&.4665$\pm.$1207  \\ 
& NODE & .0147$\pm.$0063&\textbf{.0117$\pm.$0014}& .0191$\pm.$0022&.1211$\pm.$0524&\textbf{.0114$\pm.$0019}&\textbf{.1186$\pm.$0210}\\
\cmidrule{2-8}
&  TAGODE & .0171$\pm.$0060&.0191$\pm.$0016&\textbf{.0167$\pm.$0031}&\textbf{.0790$\pm.$0792}&.0291$\pm.$0017&.1335$\pm.$0140 \\
&TAGODE-VE & \textbf{.0126$\pm.$0040}&.0125$\pm.$0015&.0193$\pm.$0029&.0849$\pm.$0727&.0160$\pm.$0014&.1297$\pm.$0221 \\
\midrule
\multirow{7}{*}{Mixed} 
& LSTM & 1.1886$\pm$1.1987&.4599$\pm.$2577&.2414$\pm.$0766&.1835$\pm.$0368&.2297$\pm.$0802&.2258$\pm.$0606\\
& GRU & 10.8797$\pm$10.4769&.4465$\pm.$2335&.5100$\pm.$2303&.2077$\pm.$0497&.2410$\pm.$0853&.3232$\pm.$0638  \\ 
& NRI &1.2360$\pm$1.4868&.7757$\pm.$0648&.2375$\pm.$0613&.2723$\pm.$3345&.3021$\pm.$1477&.3326$\pm.$1239\\
& AIDD &.4225$\pm.$0397&.3635$\pm.$0177&.3594$\pm.$0163&.3102$\pm.$5149&.2491$\pm.$0202&.3850$\pm.$0397\\ 
& NODE & .0658$\pm.$1024&.0275$\pm.$0088&.0430$\pm.$0116&.2767$\pm.$1938&.0175$\pm.$0084&.1061$\pm.$0252\\ 
\cmidrule(l){2-8}
&TAGODE &\textbf{.0430$\pm.$0619}&\textbf{.0204$\pm.$0110}& \textbf{.0180$\pm.$0083}&.2613$\pm.$4109&.0711$\pm.$0160&\textbf{.0928$\pm.$0213}
  \\
& TAGODE-VE & .0480$\pm.$0578 &.0220$\pm.$0081 & .0249$\pm.$0096&\textbf{.1126$\pm.$2042}&\textbf{.0172$\pm.$0062}&.1353$\pm.$0343 \\
\bottomrule
\end{tabular}}
\end{table*}

As shown in Table~\ref{tab:comp}, compared to the best non-continuous baseline, TAGODE achieves a maximum improvement of 94.39\% in predicting population dynamics with mixed topology, while TAGODE-VE attains a peak improvement of 95.47\% for SIS dynamics  prediction on ER networks. Our models consistently demonstrate superior performance for mixed topology regardless of dynamics, highlighting their ability to differentiate among various topology types. Compared to NODE, TAGODE reduces the error by at most 58\% for Regulatory and SIS dynamics. Note that the performance of NODE is competitive on ER and Community networks, indicating the potential to approximate these dynamics through decoupled models with mainly self-dynamics. For Population, Regulatory, and Neural dynamics, our models' predictions closely align with the ground truth, with discrepancies ranging from 0.98\% to 7.83\%.
A detailed comparison between ground truth and  predicted network states is demonstrated in Appendix~\ref{appendix_viz}.

Fig.~\ref{fig:extrap} illustrates the error evolution over the prediction window in terms of time. The model's predictions cover the time span $[0.125t_{\text{final}}, t_{\text{final}}]$, while the observation period is limited to $[0, 0.125t_{\text{final}})$. Our models demonstrate a notable capability to learn and forecast long-term trajectories. As time progresses, the error converges, reflecting the convergence of both the original and predicted dynamical systems. For SIS dynamics, TAGODE-VE outperforms TAGODE, while the latter performs slightly better on scale-free networks with population dynamics. The performance of the two ODE models is comparable for population dynamics on ER networks.
\begin{figure}[h!]
  \centering
  \includegraphics[width=0.9\linewidth]{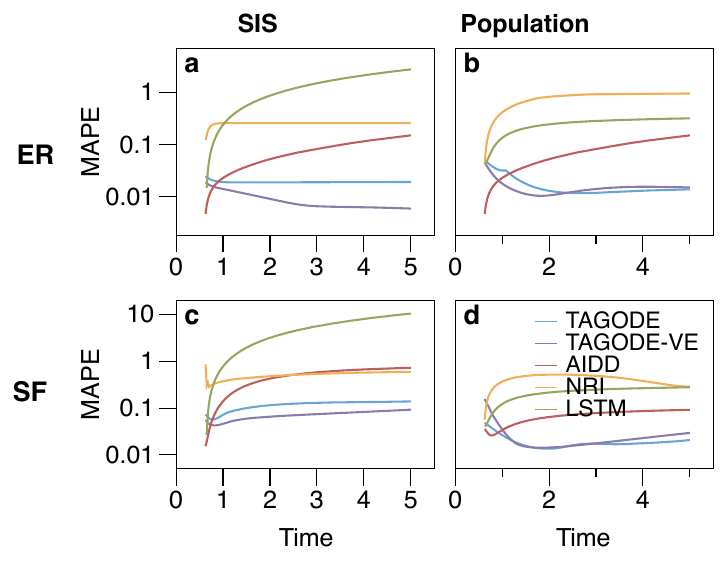}
  \caption{State inference error versus time for SIS and population dynamics on ER and SF network types. The error at each timestamp is averaged across 20 graphs and all their nodes.   \label{fig:extrap} } 
\end{figure}

\subsubsection{Out-of-distribution Study\label{sec:ood}} 
We evaluate on OOD time-series data for models trained on the mixed-topology dataset. 
In order to test the model's generalizability, we further create three OOD test datasets, featuring previously  unseen distributions of degrees  and edge weights.
The topology parameter values used to generate training and in/out-distribution test datasets are given in Table~\ref{tab:topo}. For each OOD scenario, we simulate 100 time-series datasets and report the MAPE for Mutualistic dynamics in Table~\ref{tab:comp_ood}.
TAGODE-VE surpasses all models in every OOD test case. The evolving interaction design effectively encodes additional information from the time-series data into edge embeddings compared to TAGODE, enhancing its adaptability to diverse underlying topologies. The results for the other two topologies, SF and Community, are provided in the Appendix~\ref{sec:additiona_exp} for further reference.

\begin{table}[h!]
    \centering
        \caption{Topology settings. 
        The ID (in-distribution) and OOD columns specify the parameters used to generate unweighted topology and link weights.
        The first three parameters are associated with a network type indicated in parenthesis. 
        For link weights, the  range indicates the lower and upper bound for uniform sampling.}
        %\resizebox{.95\columnwidth}{!}{%
    \begin{tabular}{lcccccc  } 
    \toprule
      {Category} &  {Parameter} & ID & {OOD}\\ 
    \midrule 
    Unweighted &$p$ (ER) & $ 0.1$ & $ 0.2 $ \\
    Topology  & $m$ (SF) & $4$ & $   8$\\ 
     & $p_{\text{out}}$  (CN) & $ 0.1$ & $ 0.2$ \\
    \midrule 
    Link Weights & - & $ [0.5,1.5]$ & $ [2,3]$\\ 
    \bottomrule
    \end{tabular} %}
    \label{tab:topo}
    \vspace{-2em}
\end{table} 

\begin{table}[h!]
\centering
\caption{MAPE in predicting trajectories of unseen sequences for Mutualistic dynamics averaged over 100 testing trajectories for each OOD scenario. The OOD test datasets draw time-series from novel unweighted topology and link weights, respectively. \label{tab:comp_ood}}
%\resizebox{\columnwidth}{!}{%
\begin{tabular}{lcc}
\toprule
 \multirow{2}{*}{Model} & \multicolumn{2}{c}{ER} \\
\cmidrule(lr){2-3}
  & TOPO & Link Weights \\
\midrule
  LSTM & .1195$\pm$.0074 & .1109$\pm$.0047 \\
  GRU & .1409$\pm$.0096 & .1311$\pm$.0071 \\ 
  NRI & .1512$\pm$.0115 & .1724$\pm$.0092 \\
  AIDD & .3060$\pm$.0207 & .3659$\pm$.0110 \\
  NODE & .1747$\pm$.0201 & .1560$\pm$.0167 \\
\cmidrule{1-3}
  TAGODE & .1799$\pm$.0125 & .2407$\pm$.0213 \\
  TAGODE-VE & \textbf{.0443$\pm$.0090} & \textbf{.0544$\pm$.0103} \\
\bottomrule
\end{tabular}%}
\vspace{-2em}
\end{table}

\vspace{-4em}
 
\section{Conclusion}
We propose a novel approach for learning network dynamics directly from time-series data, eliminating the need for prior knowledge of network topology. By observing an initial short time window, our model constructs a latent topology representation that captures the underlying system dynamics. Furthermore, the model accomodates systems with time-varying edges by modeling edge weights as functions of evolving node states, providing a unified framework for both static and dynamic networks.
This approach effectively simulates and predicts system behaviors, demonstrating robust performance in forecasting the long-term evolution of diverse dynamical systems across various complex network topologies. Our evaluation spans both transductive and inductive learning scenarios, showcasing the model’s ability to generalize across time series generated from diverse network configurations, including OOD topologies.
Future research could explore design considerations that enhance generalization to distinct network types and dynamical systems.

\section{Acknowledgement}
Y.N.D. and J.X.G. are supported by the National Science Foundation (No. 2047488), and by the Rensselaer-IBM AI Research Collaboration. 
\bibliographystyle{IEEEtranN}
\bibliography{reference}
  
\onecolumn
\appendices
\section{Experimental Setup}\label{sec:appendix_exp_setup}

\subsection{COVID-19 Dataset\label{appendix_covid}}

For COVID-19 dataset, the nodal features include  
\begin{itemize}
    \item \#Confirmed: The number of daily increased confirmed cases.
    \item \#Deaths: The number of daily increased deaths. 
    \item \#Recovered: The number of daily increased recovered cases.
    \item Testing-Rate: The number of daily cumulative test results per 100,000 persons. 
    \item Mortality-Rate: The number of daily cumulative deaths times
100 divided by the number of daily cumulative confirmed cases.
    \item Population: The of residents in each state.
    \item Mobility data: the number of people moving between points of interests (e.g., restaurants, grocery stores) within each state.
\end{itemize}
The data preprocessing procedure follows Huang et al.~\citep{huang2021coupled}. 

The training trajectory tensor is of shape $ N \times  T \times 7$, where $ N = 50, T=233 $. The training trajectory is chunked into sub-sequences of length $T_{\text{obs}}+T_{\text{pred}}$ where $T_{\text{pred}}\in \{7,14,21\}$. We set the condition length $T_{\text{obs}}$ to be 21, which is the length of the input sequence to our model. 
The data preprocessing procedure follows \cite{huang2021coupled}. Refer to Appendix~\ref{appendix_covid} for more details.

\subsection{Evaluation Metric} \label{sec:appendix_metric}
We use Mean Absolute Percentage Error (MAPE), Mean Absolute Error (MAE), and Root Mean Squared Error (RMSE) to evaluate the model performance. For trajectories with $T$ time steps, $N$ variables, and $D$ features per variable, these metric are computed as 
\begin{align}
\text{MAPE} &= \frac{1}{TND}\sum_{t=1}^{T} \sum_{i=1}^{N} \sum_{j=1}^{D} \left \lvert \frac{\hat{x}_{i,j}^{t} - x_{i,j}^{t}}{x_{i,j}^t} \right\rvert\\
\text{MAE}&= \frac{1}{TND}\sum_{t=1}^{T} \sum_{i=1}^{N} \sum_{j=1}^{D}\left \lvert \hat{x}_{i,j}^{t} - x_{i,j}^{t} \right\rvert\\
\text{RMSE}&= \sqrt{ \frac{1}{TND} \sum_{t=1}^{T} \sum_{i=1}^{N} \sum_{j=1}^{D} (\hat{x}_{i,j}^{t} - x_{i,j}^{t})^2  } 
\end{align}
where $x_{i,j}^t$ denotes the $j$th feature of the $i$th node at time $t$ and $\hat{x}_{i,j}^t$ is its corresponding prediction.

\subsection{Synthetic Network Data \label{appendix_synthetic_data}}
 
The detailed explanation of the  six  dynamics with ground truth formula is as follows.
\begin{itemize}
    \item  The susceptible-infected-susceptible model (SIS)~\citep{pastor2015epidemic,prasse2022predicting}: $\frac{d\boldsymbol{x}_i}{dt}=-\delta_i \boldsymbol{x}_i + \sum_j \boldsymbol{A}_{ij} (1-\boldsymbol{x}_i) \boldsymbol{x}_j$.   The parameter $\boldsymbol{\delta}_i > 0$ represents the rate at which individuals recover from the infection.   Following~\citep{prasse2022predicting}, we sample $\delta_i$ from 
     $$\rho(\text{diag}(1/\sqrt{\boldsymbol{\delta}^{(0)} })\boldsymbol{A}\text{diag}(1/\sqrt{\boldsymbol{\delta}^{(0)} })\boldsymbol{\delta}^{(0)}/1.5$$ with $\boldsymbol{\delta}^{(0)}\in\mathbb{R}^{N}$ and $\boldsymbol{\delta}_i\sim U[0.5,1.5]$, $\text{diag}(\cdot)$ is the $N\times N$ matrix with the input vector's values placed on the diagonal. $\rho(\cdot)$ denotes the spectral radius of the input matrix. This setting ensures the basic reproduction number~\citep{van2002reproduction} $R_0 =\rho( \text{diag} (\boldsymbol{\delta})^{ -1 }\boldsymbol{A} )$ is greater than 1, i.e., the virus does not die out.
    \item  Population dynamics:  $\frac{d \boldsymbol{x}_i } {dt} = - B \boldsymbol{x}_i^b + \sum_{j} \boldsymbol{A}_{ij} \boldsymbol{x}_j ^a$.  When $b=0$, it signifies the flow of population into and out of the locations. In contrast, when $b>0$, it corresponds to mortality within the population. For our analysis, we have set $B=1$, $b=0.5$, and $a=0.2$.
    \item Gene regulatory dynamics: $\frac{d \boldsymbol{x}_i}{dt} = - B_i \boldsymbol{x}_i^f + \sum_{j} \boldsymbol{A}_{ij} \frac{\boldsymbol{x}_j^2}{\boldsymbol{x}_j^2+1}$~\citep{alon2006introduction,gao2016universal}.  The parameter $B_i$ controls the rate of self-decay, where $f=1$ corresponds to degradation, and $f=2$ corresponds to dimerization, a process by which two identical molecules come together to form a dimer. Additionally, $h\geq 0$ represents the Hill coefficient, which quantifies the rate of saturation for the impact of neighboring nodes.   For our analysis, we set the values as ($B_i=1$, $f=1$, $h=2$).  
    \item Mutualistic dynamics:    $\frac{d\boldsymbol{x}_i}{dt}=B_i \boldsymbol{x}_i (1-\frac{\boldsymbol{x}_i^{a}}{C_i})+\sum_{j} \boldsymbol{A}_{ij} \boldsymbol{x}_i \alpha\boldsymbol{x}_j^h (1+\alpha \boldsymbol{x}_j^h)^{-1}$~\citep{hens2019spatiotemporal}. The first term resembles logistic growth, where $B_i$ signifies the reproduction rate, $a$ determines intraspecific competition, and $C_i$ represents the carrying capacity. The mutualistic aspect is embodied in the saturating term $\alpha\boldsymbol{x}_j^h \left(1+\alpha \boldsymbol{x}_j^h\right)^{-1}$, where $h$ controls the rate of saturation. We set $\alpha=B_i=C_i=1$ and consider cubic growth restriction $(a=2)$ with a  neighbor cooperation level of $h=1$.  
    \item Neural dynamics~\citep{wilson1972excitatory}: $\frac{d\boldsymbol{x}_i}{dt}=-\boldsymbol{x}_i+ \sum_j \boldsymbol{A}_{ij} (1+\exp(-\tau(\boldsymbol{x}_j -\mu)))^{-1}$.  The parameters $\tau$ and $\mu$ determine the slope and threshold of the neural activation function, respectively.   
    \item The Lotka–Volterra model (LV)~\citep{macarthur1970species}: $\frac{d\boldsymbol{x}_i}{dt} = \boldsymbol{x}_i (\alpha_i - \theta_i \boldsymbol{x}_i ) - \sum_j \boldsymbol{A}_{ij} \boldsymbol{x}_i \boldsymbol{x}_j $.  The dynamical parameters govern the growth of species $i$ with $\alpha_i, \theta_i >0$. In the experiments, we sample $\alpha_i,\theta_i$ from a uniform distribution in the range $[0.5,1.5]$.
\end{itemize}
Note that dynamics other than Lotka-Volterra are cooperative, i.e., the increase of neighbor activity stimulates the growth of the current node ($\frac{\partial g}{\partial \boldsymbol{x}_j} \geq 0$).
The ground truth time-series data is simulated by solving the ODEs given initial conditions (Table~\ref{tab:ic}) and dynamical parameters.

\begin{table}[h]
    \centering
    %\resizebox{.95\columnwidth}{!}{
        \caption{ Initial conditions and final time used to simulate the time-series data. Initial conditions include uniform (ID) and normal   distributions $\mathcal{N}(\mu,\sigma )$ with mean $\mu$ and standard deviation $\sigma$ (OOD). Ground truth trajectories comprise 200 timestamps in $[0, t_{\text{final}}]$. For SIS, we initially set half of the nodes to be infected with $x_i \geq 0.5$ and half to be uninfected $x_i<0.5$.}
    \begin{tabular}{lcccc  }
    \toprule 
    \multirow{2}{*}{Dynamics} & \multicolumn{2}{c}{Initial Condition }    & \multirow{2}{*}{$t_{\text{final}}$}\\
    & ID & OOD &\\ 
    \midrule
    SIS & $ \{0.1,0.8\}$   & $\mathcal{N}( 0.5, 0.1)$  & 5\\    
    Population & $U(0,2)$   & $\mathcal{N}( 6, 1)$ & 5\\    
    Regulatory & $U(0,2)$ & $\mathcal{N}( 6,  .1)$& 5\\  
    Mutualistic  & $U(0,5)$&  $\mathcal{N}( 6, 1)$ & 2 \\ 
    Neural &  $U(0,10)$ & $\mathcal{N}(   6, 1)$&  5\\   
    LV &  $U(0,20)$& $\mathcal{N}(  6, 1)$&   5\\  
    \bottomrule
    \end{tabular} 
    \label{tab:ic}
\end{table}  
 
For synthetic networks, the adjacency matrix is determined by the distribution of node degrees and link weights. We explore three distinct types of network models, each representing a unique process of network generation.
(1) Erd\H{o}s-R\'enyi (ER) networks~\citep{erdds1959random}, also known as random networks, where node degrees are drawn from a Poisson distribution with an average degree  $\langle \boldsymbol{k}\rangle=(N-1)p$. ($p$ is the probability of edge creation).
(2) Scale-free (SF) networks~\citep{barabasi1999emergence}, characterized by a Power-law degree distribution. We apply  preferential attachment  to construct this network topology. We begin with a set of $m_0$ nodes and   iteratively introduce new nodes. Each new node is connected to $m$ existing ones, with the likelihood of connecting to an existing node being  proportional to that node's degree. 
(3) Community networks~\citep{fortunato2010community}, generated by the random partition graph models with intra-cluster connection of $p_{\text{in}}$ and inter-cluster connection of $p_{\text{out}}$.  In our experiment, we fix the value of $p_{\text{in}}$ at 0.25. When simulating network structures, we first generate an unweighted graph using the network model and assign each link with a weight 
drawn uniformly from the range of $[0.5, 1.5]$ to form the ground truth topology $A$.

For each dynamics paired with a topology type, we generate 140 realizations of networks with 100 nodes and sample 100/20/20 sequences for train/val/test, respectively. We combine different topology into a mixed dataset with 300/60/60 train/val/test sequences for every dynamics. We report results for models that are trained and tested solely on one network topology type as well as the mixed dataset. 
The performance is quantified by MAPE.  For Lotka-Volterra dynamics, we employ RMSE due to the predominance of states near zero.  
The observational sequence length $T_{\text{obs}}$ is set to 25, and the prediction window contains 175 time steps in both training and testing.

\section{Additional Experiments\label{sec:additiona_exp}}

For all experiments, we adopted AdamW optimizer and a cosine annealing learning rate scheduler with starting value 0.01 and a maximal epoch of 80. 
The hidden dimension for all models was set to 20 for \textbf{COVID-19} dataset and 16 for other experiments.  
The number of attention heads applied for our time-varying edge ODE   is 1 for SIS on ER and SF networks and 3 for other combinations of dynamics and topology. For NRI, \# edge types was set to 3 for \textbf{COVID-19} dataset. In other cases, \# edge type was set to 1.

\subsection{Scalability \label{appendix_scalability}}

In the test stage, the model can be applied to larger networks with similar degree distributions.
We conducted an extrapolation task on an ER network (N=10,000, Avg. degree=10), using a model trained on smaller ER networks (N=100, Avg. degree=12), resulting in a MAPE of 0.026 over the prediction window (Table~\ref{tab:size-generaliation}). Due to memory constraints with NRI and the inability of architectures like LSTM and AIDD to adapt to networks of varying sizes, we don't have their respective results. Additionally, we tested two pre-trained models with evolving latent edges on a real protein-protein interaction topology~\citep{agrawal2018large} with 21,557 nodes and 342,353 edges. The nodal states are in the range [0.00172, 2035.58], with a mean of 27.58 and a standard deviation of 59.84. The following table reports the median absolute percentage error.

\begin{table}[ht]
    \centering
    %\resizebox{.95\columnwidth}{!}{
        \caption{ Train on small networks and test on larger networks. }
    \begin{tabular}{lccc  } 
    \toprule
    \multirow{2}{*}{Pred Length} & \multicolumn{2}{c}{Pre-trained Topology}         \\
   & ER 
& SF    \\
\midrule
10                & 0.107                & 0.44  \\
50                & 2.0121               & 0.744 \\
100               & 0.989                & 1.766 \\
175               & 0.671                & 2.47 \\
    \bottomrule
    \end{tabular} 
    \label{tab:size-generaliation}
\end{table}

\subsection{Additional OOD Study}

\subsubsection{Complete Results for All Topology types}
Table~\ref{tab:comp_ood_app} presents the results for all topology types under the same OOD setting described in Section~\ref{sec:ood}.
\begin{table*}[h!]
\centering
\caption{MAPE in predicting trajectories of unseen sequences for Mutualistic dynamics averaged over 100 testing trajectories for each OOD scenario.  The OOD test datasets draw time-series from novel  unweighted topology, and link weights, respectively.   \label{tab:comp_ood_app} }
\resizebox{\textwidth}{!}{%
\begin{tabular}{lcccccc}
\toprule
 \multirow{2}{*}{Model} & \multicolumn{2}{c}{ER} & \multicolumn{2}{c}{SF} & \multicolumn{2}{c}{Community} \\
\cmidrule(lr){2-3} \cmidrule(lr){4-5} \cmidrule(lr){6-7} 
  & TOPO & Link Weights & TOPO & Link Weights & TOPO & Link Weights \\
\midrule
  LSTM &  .1195$\pm.$0074&.1109$\pm.$0047&.1476$\pm.$0112&.1371$\pm.$0093&.1170$\pm.$0061&.0961$\pm.$0026 \\
  GRU &   .1409$\pm.$0096&.1311$\pm.$0071&.1697$\pm.$0121&.1604$\pm.$0106&.1383$\pm.$0085&.1137$\pm.$0052\\ 
  NRI &  .1512$\pm.$0115&.1724$\pm.$0092&.1254$\pm.$0370&.1454$\pm.$0191&.1584$\pm.$0220&.2110$\pm.$0052\\
  AIDD &   .3060$\pm.$0207&.3659$\pm.$0110&.2406$\pm.$1075&.2614$\pm.$0539&.3342$\pm.$0575&.4734$\pm.$0096\\
  NODE &  .1747$\pm.$0201&.1560$\pm.$0167&.2125$\pm.$0260&.1948$\pm.$0188&.1680$\pm.$0203&.1377$\pm.$0105\\
  \cmidrule{1-7}
   TAGODE &  .1799$\pm.$0125&.2407$\pm.$0213&.1662$\pm.$0321&.2032$\pm.$0191&.1994$\pm.$0218&.3758$\pm.$0189\\
   TAGODE-VE &  \textbf{.0443$\pm.$0090}&\textbf{.0544$\pm.$0103}&\textbf{.0550$\pm.$0194}&\textbf{.1087$\pm.$0490}&\textbf{.0503$\pm.$0118}&\textbf{.0633$\pm.$0116}\\
\bottomrule
\end{tabular}}
\end{table*}

\subsubsection{Generalization across density and network types}
\begin{figure}[h]
  \centering 
  \begin{subfigure}[b]{0.4\textwidth}
  \includegraphics[width=\linewidth]{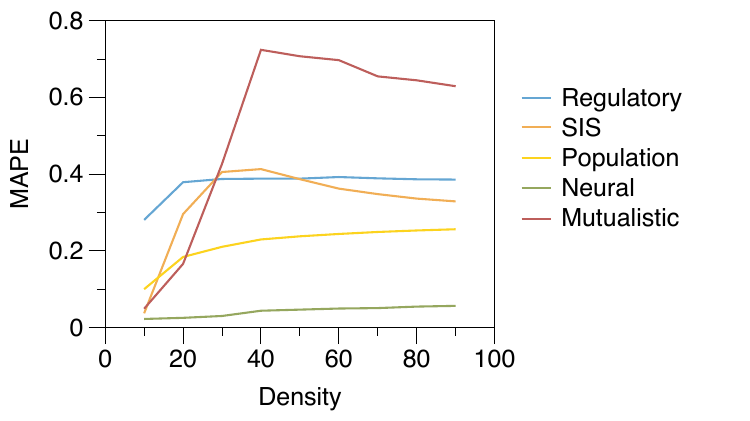}
  \caption{Generalization across density.   \label{fig:ood_density}} 
  \end{subfigure}% 
  \begin{subfigure}[b]{0.6\textwidth}
  \centering
  \includegraphics[width=\linewidth]{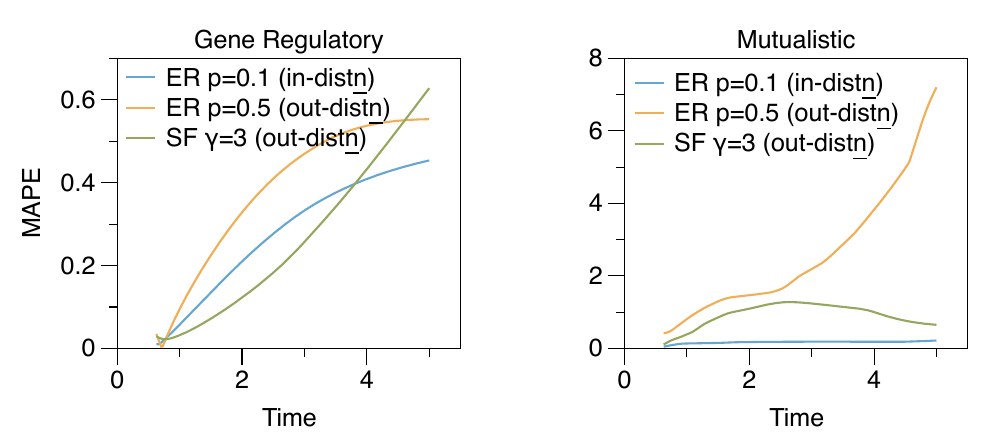}
  \caption{ Generalization across network types.  \label{fig:ood_topology}} 
\end{subfigure}
\caption{OOD test cases}
\end{figure}

Fixing the network density at training at 0.1.
We increase the density ($\langle \boldsymbol{k}/N$) from 0.1 to 0.9 and plot the average MAPE in Fig.~\ref{fig:ood_density}. For most dynamics, the performance is stable as density varies. 
Fig.~\ref{fig:ood_topology} further demonstrates the model's ability to generalize to unseen distribution. 

\subsubsection{Comparison with Baselines}
We create a dataset of 96 time-series generated on ER networks with 100 nodes and average degree 10 for each type of dynamics. The link weights are drawn randomly from a uniform distribution $U[0.5, 1.5]$. During the training phase, we utilize 76 sequences, reserving 20 for testing. We introduce 20 sequences with novel distributions related to initial conditions, network topology, and edge weights. The conditioning window is set to 25 time steps, and the prediction window is 100 time steps.   In testing, the conditioning window remains at 25 time steps, while the prediction length extends to 175 time steps.

When faced with unseen time-series, the most substantial drop in performance occurs when perturbations are introduced to the distribution of link weights  (Table~\ref{tab:comp_ood1}).
In cases involving out-of-distribution  initial conditions, the NRI model surpasses our model for both Mutualistic models and SIS epidemics. Additionally, when confronted with out-of-distribution link weights, the LSTM model outperforms our model for dynamics including Regulatory, and SIS. When presented with unseen unweighted topologies, our model maintains more competitive over other baselines.  

\begin{table}[h!]
\centering
\caption{Error in predicting trajectories of unseen sequences. The in-distribution (ID) test dataset consists of time-series data generated from the same distribution of initial conditions (IC), unweighted topology (TOPO), and link weights. The OOD test datasets draw time-series from unseen initial condition, unweighted topology, and link weights, respectively.  We report RMSE for Lotka-Volterra and MAPE for other dynamics.  }
  \label{tab:comp_ood1}
  \begin{tabular}{lllllll}
    \toprule
    \multirow{2}{*}{Dynamics} &\multirow{2}{*}{Method}  
    & \multirow{2}{*}{ID} & \multicolumn{3}{c}{OOD}   \\
    & & & IC  & TOPO & Weights
      \\ 
    \midrule 
  \multirow{3}{*}{SIS}   
  &NRI  &  0.207 & \textbf{0.164}&0.176 &  \textbf{0.15}\\
    & LSTM & 0.154 &0.369 &0.148 & 0.314 \\ 
    &TAGODE-VE& \textbf{0.051}  &0.340   &\textbf{0.041}  & 0.367\\ 
    \midrule 
    \multirow{3}{*}{Population} 
    &NRI   &  0.198 & 0.184& 0.238&0.624\\
    & LSTM& 0.214 & 0.234&0.282 &0.633\\
    &TAGODE-VE& \textbf{0.019} & \textbf{0.028} &\textbf{0.11}   & \textbf{0.302 }\\  
    \midrule 
   \multirow{3}{*}{Regulatory}  
   & NRI & 0.724  &0.652 &0.675 &0.897\\
   & LSTM & 0.199 &0.325 &0.195 & \textbf{0.339}\\  
   &  TAGODE-VE& \textbf{0.033} & \textbf{0.194} & \textbf{0.146} &  0.758\\ 
     \midrule 
    \multirow{3}{*}{Mutualistic} 
    &NRI &  0.109   & \textbf{0.051}&0.102 &0.447\\
    &LSTM & 0.178 &0.222 & 0.173& 0.095\\ 
    &TAGODE-VE& \textbf{0.081} & 0.125 &\textbf{0.038}  & \textbf{0.059} \\  
    \midrule
    \multirow{3}{*}{Neural}&NRI&  0.657 &0.653 &  0.646&629.29\\
    &LSTM & 0.179 &0.16 &0.17 &0.332\\ 
    &TAGODE-VE & \textbf{0.018}  &\textbf{0.023} &\textbf{0.017}& \textbf{0.116}\\ 
     \midrule
    \multirow{3}{*}{Lotka-Volterra}
    &NRI& 0.531 & 0.373 & 0.59 & 0.477 \\
    &LSTM & 0.175 & 0.147 & 0.161 & 0.179 \\ 
    &TAGODE-VE  & \textbf{0.085}& \textbf{0.071} &\textbf{0.057}&\textbf{0.171} \\ 
    \bottomrule 
\end{tabular}  
\end{table} 
\subsection{Hyperparameter Study}
\begin{figure}[h]
  \centering
  \includegraphics[width=0.5\linewidth]{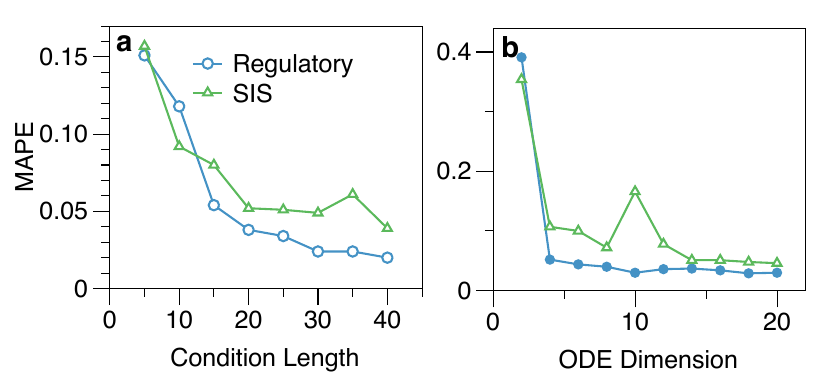}
  \caption{Hyperparameter study: condition length and latent dimension.\label{fig:hyper} } 
\end{figure}

We investigate two hyperparameters on the impact of the static-edge ODE model: condition length and latent ODE dimension. To prevent excessive observation, we set a maximum condition length of less than $0.2t_{\text{final}}$. Figure~\ref{fig:hyper} illustrates that inference accuracy improves as the model observes more data points to compute the latent initial condition. Additionally, as the ODE dimension increases, performance improves and eventually plateaus.

\subsection{Robustness}
We investigate the impact of observational noise by applying a Gaussian distributed noise with mean $x_i$ and standard deviation $\sigma \lvert x_i\rvert$ to the condition window. The training data is the same as above and use 20 ER networks drawn from identical distribution for testing. The error growth is bounded with $O(\sigma)$ for Regulatory and Neural dynamics.
\begin{figure}[h]
  \centering
  \includegraphics[width=0.4\linewidth]{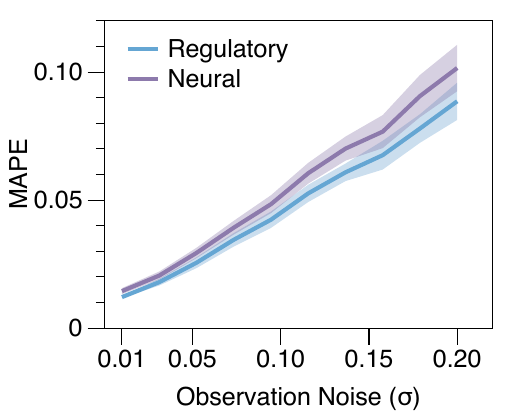}
  \caption{Effect of observation noise on the condition window. \label{fig:noise}} 
\end{figure}
\subsection{Encoder Variants Comparison\label{sec:exp_encoder}}
We study four different types of encoders for node embeddings.  Let $\boldsymbol{x}_i \in \mathbb{R}^{T_{\text{obs}}D }$ denote the feature vector of node $i$ concatenating observations at each timestamp.  
\begin{itemize}
    \item FeedForward (FFW). Node embeddings are acquired through a two-layer Multilayer Perceptron (MLP). Subsequently, the latent pairwise interaction vector is determined via a feedforward function applied to the concatenation of node embeddings:
    \begin{align}
    \boldsymbol{z}_i(0) &= f_{\text{node}}(\boldsymbol{x}_i) 
    \end{align}
    \item NRI-based. In addition to FFW, the NRI-based encoder incorporates an extra round of message passing from edges to nodes for node embedding updates:
    \begin{align}
    \boldsymbol{h}_i  &= f_{\text{node}}(\boldsymbol{x}_i)\\
    \boldsymbol{h}_{j\rightarrow i}  &= f_{\text{edge}}^1 ([\boldsymbol{h}_j || \boldsymbol{h}_i])\\
    \boldsymbol{z}_i(0)&= f_{\text{edge2node}} (\sum_{j\neq i} \boldsymbol{h}_{j \rightarrow i}) 
    \end{align} 
    \item Graph Transformers (GT). Initially, input is mapped to a latent space, followed by computation of attention scores for each pair of latent nodal embeddings. The final node vector is generated by aggregating the weighted message from each node, along with a residual connection from the node itself. Lastly, the connection strength embedding is derived by feeding the concatenation of node embeddings to an MLP:
    \begin{align}
    \boldsymbol{h}_i  &= f_{\text{node}}(\boldsymbol{x}_i)\\  
    e_{ij}&= \text{LeakyReLU}(  (\boldsymbol{W}_{k}\boldsymbol{h}_j )^{\top} (\boldsymbol{W}_q \boldsymbol{h}_i) )\\
    \alpha_{ij} &= \text{softmax} (e_{ij})\\
    \boldsymbol{z}_i(0)&= \boldsymbol{h}_i + \sigma ( \sum_{j\in \mathcal{V}} \alpha_{ij} \boldsymbol{W}_v \boldsymbol{h}_j  ) 
    \end{align}
    \item Graph Transformers on Dynamic Graphs (GT-DG)~\citep{huang2021coupled}. Assuming a fully connected graph, we construct a dynamic graph where each node represents an observation at a specific timestamp $\boldsymbol{x}_i^t$. Edges are formed between $\boldsymbol{x}_i^{t_1}$ and $\boldsymbol{x}_j^{t_2}$ if $i\neq j$ and $t_1=t_2$ or $i=j$ and $t_1<t_2$. This dynamic graph is then processed through the previously defined graph transformer to obtain both node and edge embeddings.
\end{itemize}
We implement the above functions $f_{\cdot}$ as FeedForward networks
\begin{align}
f_{\cdot}(\boldsymbol{h})=
\mathrm{FeedForward}(\boldsymbol{h}) 
    =  \boldsymbol{W}^{(2)}
\sigma(\boldsymbol{W}^{(1)}\boldsymbol{h}  + \boldsymbol{b}^{(1)}) +\boldsymbol{b}^{(2)} 
\end{align}
We adopt an instance of the neural ODE with time-varying edges:
\begin{align}
\frac{d\boldsymbol{z}_i (t)}{dt} &=
  \boldsymbol{W}\left(   \sum_{j=1}^{N} \overset{3}{\underset{h=1}{\parallel}} 
    -\boldsymbol{z}_i  +\boldsymbol{\hat{A}}_{ij}^{h} \boldsymbol{W}_v^h\boldsymbol{z}_j(t)    \right)+  \boldsymbol{b}\label{eq:node3}
\end{align}   
where $\boldsymbol{W}\in\mathbb{R}^{d\times 3d}$ projects the concatenation interaction terms to the latent space.

We simulate 96 ER network time-series data with $N=100$ and average degree 10 and use 76 sequences for training, and 20 for testing.  
In Table~\ref{tab:encoder}, we present a comparison between the ground truth and predicted trajectories, showcasing the activities of 20 nodes over varying time steps.

Table~\ref{tab:encoder} shows that the performance degrades from FFW, GT, GT-DG to NRI. The relative worse performance for GT-DG can be attributed to the lack of ground truth topology.  
While both GT and NRI consider the interaction between nodes, the latter has a worse performance.  

The NRI-based and GT-DG models encounter difficulties in distinguishing unique behaviors among individual nodes, often collapsing multiple nodes' trajectories into a single representative path. 

\begin{table}[h]
\centering
\caption{State reconstruction error for different encoder variants. (RMSE for LV and MAPE for other dynamics).\label{tab:encoder}}
\begin{tabular}{lcccc}
\toprule
\textbf{Dynamics} & \textbf{FFW} & \textbf{NRI} & \textbf{GT} & \textbf{GT-DG} \\
\midrule
SIS & \textbf{0.051} & 0.322 & 0.060 & 0.125 \\
Population & \textbf{0.019} & 0.324 & \textbf{0.019} & 0.322 \\
Regulatory & \textbf{0.033} & 0.285 & 0.040 & 0.303 \\
Eco2 &\textbf{0.081} & 0.289 & 0.095 & 0.206 \\
WC & \textbf{0.018} & 0.282 & 0.041 & 0.161 \\
LV & \textbf{0.081} & 0.289 & 0.088 & 0.143 \\
\bottomrule
\end{tabular} 
\end{table}

\section{Network State Visualization \label{appendix_viz}}
We visualize the true and predicted states  following~\citep{zang2020neural}.  For each network, we order the 100 nodes in ascending order of final state, and plot the value of node $i$ at coordinate $(\lfloor i/\sqrt{10}\rfloor,i\mod{\sqrt{10}})$.  
 
\begin{figure*}[t]
  \centering
  \includegraphics[width=0.9\linewidth]{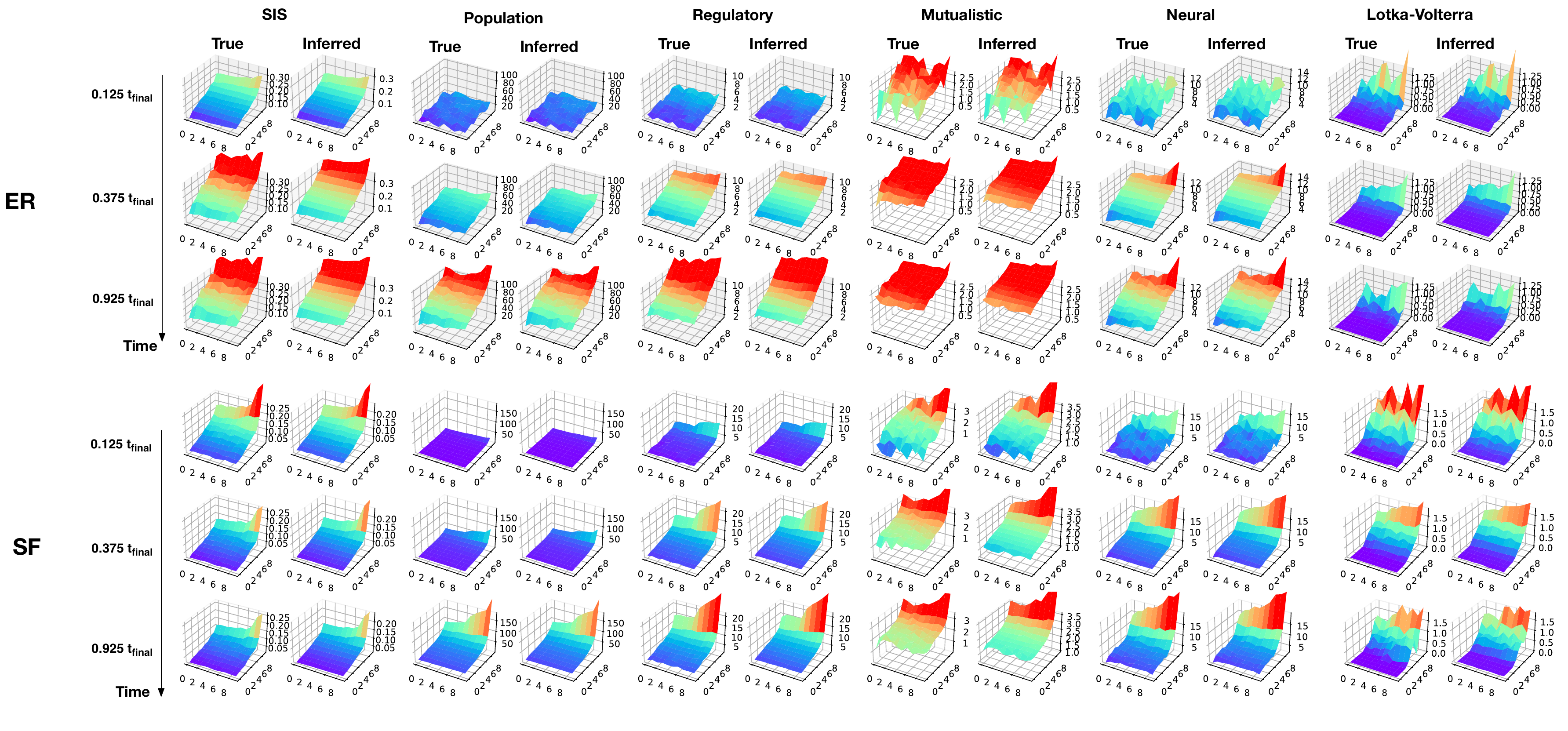} 
  \includegraphics[width=\linewidth]{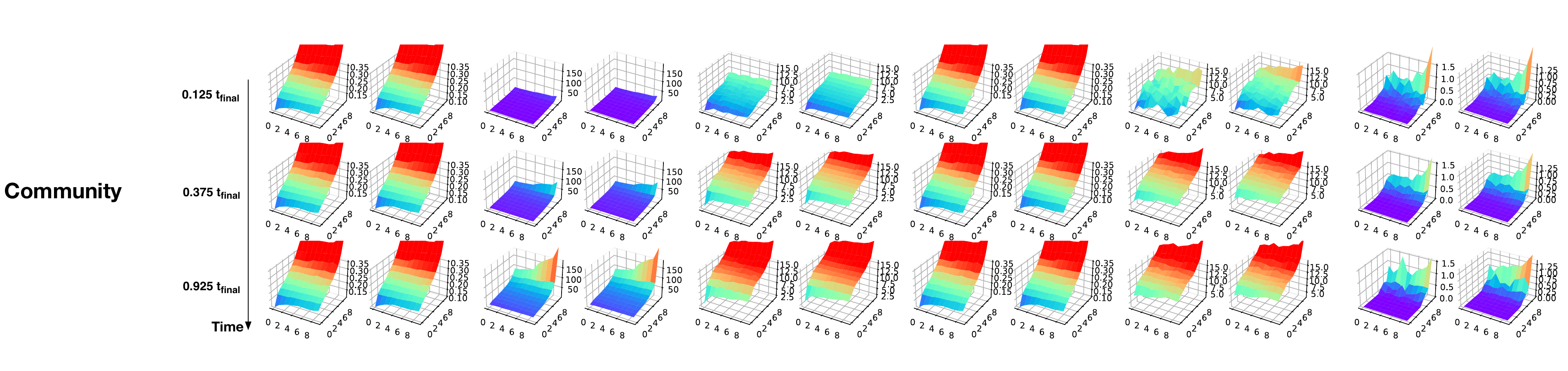}
  \caption{ Comparison of ground truth and predicted states at three discrete time steps $0.125t_{\text{final}}, 0.375t_{\text{final}},0.925t_{\text{final}}$. The node index is sorted according to ascending order of final states. We represent the system states on a $\sqrt{N}\times \sqrt{N}$ grid, where each node $i$'s value is displayed at coordinates $(\lfloor i/\sqrt{N}\rfloor,i\mod{\sqrt{N}})$\label{fig:3d_state_com}. } 
\end{figure*}

\end{document}